\documentclass[onecolumn,11pt]{article}
\usepackage[top=1in, bottom=1in, left=1in, right=1in]{geometry}
\usepackage{amsmath,amsfonts,amscd,amssymb}
\usepackage{graphicx}
\usepackage{epstopdf}
\usepackage{overpic}
\usepackage{cancel}
\usepackage{rotating}
\usepackage{url}
\usepackage{caption}
\usepackage{color}
\usepackage{rotating}
\usepackage{multirow}
\usepackage{wrapfig}
\usepackage{mathtools}
\usepackage{subeqnarray}
\usepackage{setspace}
\usepackage{palatino} 
\usepackage[numbers,sort&compress]{natbib}

\usepackage[bottom,flushmargin,hang,multiple]{footmisc}
\usepackage{lipsum}
\newcommand\blfootnote[1]{%
  \begingroup
  \renewcommand\thefootnote{}\footnote{#1}%
  \addtocounter{footnote}{-1}%
  \endgroup
}

\usepackage[utf8]{inputenc} 
\usepackage[T1]{fontenc}    
\usepackage{hyperref}       
\usepackage{url}            
\usepackage{booktabs}       
\usepackage{amsfonts}       
\usepackage{nicefrac}       
\usepackage{microtype}      

\usepackage{graphicx}
\usepackage{amsmath}
\usepackage{amsthm}
\usepackage{amssymb}
\usepackage{mathtools}
\usepackage{color}

\usepackage{times}
\usepackage{graphicx} 
\usepackage{subfigure}

\usepackage{algorithm}
\usepackage{algorithmic}

\usepackage{hyperref}

\usepackage{times}
\usepackage{xcolor}
\usepackage{soul}
\usepackage[utf8]{inputenc}

\usepackage{latexsym}
\usepackage{booktabs}
\usepackage{amsmath}
\usepackage{verbatim}
\usepackage{amssymb}
\usepackage{amsthm}
\usepackage{algorithm}
\usepackage{algorithmic}
\usepackage{graphicx}
\usepackage{subfigure}
\usepackage{booktabs}
\usepackage{epstopdf}
\usepackage{lipsum}
\usepackage{color}

\usepackage[utf8]{inputenc} 
\usepackage[T1]{fontenc}    
\usepackage{hyperref}       
\usepackage{url}            
\usepackage{booktabs}       
\usepackage{amsfonts}       
\usepackage{nicefrac}       
\usepackage{microtype}      

\usepackage{multicol}

\usepackage{hyperref}
\usepackage{url}

\newcommand{\shrink}[1]{}

\usepackage{wrapfig}

\title{Learning  Brain Dynamics from Calcium Imaging with Coupled van der Pol and LSTM}

\author{\normalsize{German Abrevaya$^{1*}$, Irina Rish$^2$, Aleksandr Y.~Aravkin$^3$, Guillermo Cecchi$^{2}$, James Kozloski$^2$},\\ 
\normalsize{ Pablo Polosecki$^2$, Peng Zheng$^3$, Silvina Ponce Dawson$^1$, Juliana Rhee$^4$, David Cox$^2$}\\
\footnotesize{$^1$ Departamento de Física, FCEyN, UBA and IFIBA, CONICET, 1428 Buenos Aires, Argentina}\\
\footnotesize{$^2$ IBM T.J. Watson Research Center, Yorktown Heights, New York, 10598, USA}\\
\footnotesize{$^3$ University of Washington, Seattle, WA 98195, USA}\\
\footnotesize{$^4$ Harvard University, Cambridge MA, USA}
}
\date{}
\begin{document}
\maketitle

\blfootnote{$^*$ Corresponding author (kpchamp@uw.edu).}
\begin{abstract} 
Many real-world data sets, especially in biology, are produced by complex  nonlinear dynamical systems. In this paper,  we focus on brain calcium imaging (CaI)  of different organisms (zebrafish and rat), aiming to build a model of joint activation dynamics in large neuronal populations, including the whole brain of zebrafish.
We propose a new approach for capturing dynamics of temporal SVD components that uses the coupled (multivariate) van der Pol (VDP) oscillator, a nonlinear  ordinary differential equation (ODE) model describing neural activity, with a new parameter estimation technique that combines variable projection optimization and stochastic search.  We show that the approach successfully handles nonlinearities and hidden state variables in the coupled VDP.  The approach  is accurate, achieving 0.82 to 0.94 correlation between the  actual  and model-generated components, and interpretable, as VDP's coupling matrix reveals anatomically meaningful positive (excitatory) and negative (inhibitory) interactions across different brain subsystems corresponding to spatial SVD components. Moreover,  VDP is comparable to (or sometimes better than)  recurrent neural networks  (LSTM) for (short-term) prediction of future brain activity; VDP needs less parameters to train, which was a plus on our small training data.  Finally, the overall best predictive method, greatly outperforming  both VDP and LSTM in short- and long-term predicitve settings on both datasets, was the new  hybrid VDP-LSTM approach that used VDP to simulate large domain-specific dataset for LSTM pretraining; note that simple LSTM data-augmentation via  noisy versions of training data was much less effective.

\end{abstract}

\section{Introduction}

 Neuroscience is a source of exceedingly interesting data for analysis and modeling, with examples ranging from connectome structural information ~\cite{takemuraNature2013}, to multi-electrode recordings~\cite{Schwartz2014} and functional magnetic resonance 4D imaging ~\cite{Feinberg2010}. These experimental techniques, however, have intrinsic resolution limitations that preclude a closer connection between signal analysis and neurophysiology-based modeling. A recently introduced technique, brain-wide calcium imaging (CaI) ~\cite{ahrensNature2012}, provides a unique perspective on neural function, recording the concentrations of calcium at sub-cellular spatial resolution across an entire vertabrate brain, and at a temporal resolution that is commensurable with the timescale of calcium dynamics~\cite{ahrensNatureMethods2013}. Furthermore, recovery of action potential timing and circuit structure and dynamics using this modality of recording has an established history in the field ~\cite{peterlinPNAS2000}, but not at whole brain scales. 
 Therefore, {\em this new source of CaI data poses the challenge of how to analyze calcium signals derived from these methods in order to obtain scientific insights about whole brain function\footnote{So far,  most of the recent work on calcium imaging is  primarily focused on  identifying the sources (neurons and axonal or dendritic
processes) and  denoising/deconvolving the neural activity from calcium signals \cite{Apthorpe_NIPS2016,robust-estimation-of-neural-signals-in-calcium-imaging,Speiser_NIPS2017,giovannucci2017onacid}.}.}

Linear vector-autoregressive (VAR) models may not be able to capture compex nonlinear dynamics (as confirmed by our experiments), while popular nonlinear general-purpose temporal models such   recurrent neural networks (RNNs) are not easily interpretable from neuroscience perspective, i.e. do not immediately reveal functional connectivity across different brain subsystems (spatial components) from  observed coupled dynamics; moreover, they tend to require much larger amounts of training data than typically available in neuroimaging.

Thus, our motivation here was to propose an interpretable,   domain-specific nonlinear dynamical model able to capture the most relevant features of a complex nonlinear dynamical system, such as brain activity in neuroimaging data.  Brain activity exhibits a highly nonlinear behavior that can be oscillatory or even chaotic~\cite{korn2003there}, with sharp phase transitions between different states.  The simplest models that can capture these behaviors are {\it relaxation oscillators}.  One of the most famous examples is the van der Pol  (VDP) oscillator~\cite{guckenheimer2013nonlinear}, used to
model a variety of problems in physics. It is also highly relevant to neuroscience, as a special case of the FitzHugh-Nagumo model, which is in turn  a simplification  of the classical
 Hodgkin–Huxley model of activation and deactivation dynamics of a spiking neuron ~\cite{Izhikevich2007,fitzhugh1961impulses}.  VDP  can also model  neural populations through its approximation of Wilson-Cowan dynamics~\cite{kawahara1980, destexhe2009}. Moreover, under generic assumptions of neuronal connectivity, it has been shown to be the {\sl simplest} nonlinear model that sustains a meta-stable center manifold such that brain activity is poised at the boundary between stability and instability, a condition that allows for significant coding advantages~\cite{moirogiannis2017renormalization, alonso2019, seung2000}.
 
 However, estimating both states and structure parameters of the   coupled VDP model from noisy data, with model nonlinearities and unobserved variables, and  can be nontrivial, as discussed in the next section.  We propose an efficient local technique for an inexact (simplified) VDP formulation, which combines variable-projection optimization and stochastic search in order to get out of potential local minima and "jump" to more promising parts of an enormous search space. Our approach works well empirically on two CaI datasets, accurately fitting the data; furthermore, VDP coupling matrix can be used to analyze positive and negative  interactions  across different brain areas (spatial components) to obtain neurocientific  insights.
 However, an in-depth evaluation of the approach on various simulated data, other dynamical systems, as well as comparison with other methods (among those applicable to our problem) is beyond the scope of this paper, and remains the direction for future work.

Finally, we also use VDP model for predicting the future brain activity, and obtain good short-term prediction results, clearly outperforming VAR model and comparable to/better than LSTM. However, the  best predictive accuracy is achieved a hybrid method, which uses   VDP learned  from relatively limited training data to simulate large amounts of pre-training data for LSTM, pulling it towards specific nonlinear dynamics, and then fine-tuning it on limited-size  real data.  We demonstrate that this hybrid approach  consistently improves LSTM performance, on both datasets, and both for short- and long-term predictions. Moreover, data-augmentation with VDP simulations boosts LSTM performance significantly higher than just using   noisy versions of training data.

\section{Related Work}

Estimating  states and parameters of (multivariate) ordinary differential equations (ODEs) from data is a problem of significant interest in  various research communities, ranging from physics and biology to control theory, machine learning and optimization. However, this problem can be quite challenging, due to the high computational cost of numerical integration involved in evaluation of each candidate set of parameters, and large search spaces, especially for high-dimensional ODEs. These difficulties have inspired a range of different approach to the problem. Many traditional methods  avoid optimization by using the unscented Kalman filter ~\cite{quach2007estimating,voss2004nonlinear,sitz2002estimation} or other derivative-free dynamic inference methods~\cite{havlicek2011dynamic}. However, derivative-free methods have limitations --  there is no convergence criteria or  disciplined way to iterate them to improve estimates. 
Optimization-based approaches for fitting parameters and dynamics are discussed by~\cite{gabor2015robust}, who formulate parameter identification under dynamic constraints as an ODE-constrained optimization problem. We take a similar view, and use recent insights into variable projection to develop an efficient optimization algorithm
for  state  and parameters estimation of the van der Pol (VDP), where state vector includes unobserved (hidden) variables.  The work of~\cite{gabor2015robust} is focused on global strategies (e.g. multiple re-starts of well-known methods); our contribution is to develop an efficient local technique for an inexact VDP formulation. 

There is also a growing family of   {\em gradient matching techniques} \cite{babtie2014topological,macdonald2015gradient}
based on minimizing the difference between the interpolated slopes and the time derivatives of the state variables in the ODE’s, and extending earlier work on spline-based methods \cite{varah1982spline,ramsay2007parameter}. However, it is difficult to adapt spline-based methods for learning ODEs with unobserved variables (as in case of VDP model considered here); similar limitations affect recently proposed Gaussian process   regression approaches \cite{calderhead2009accelerating,dondelinger2013ode}. In \cite{gorbach2017scalable}, a variational inference approach is proposed which is able to handle unobserved states; however, the method only considers ODEs  which are locally linear in parameters and states -- the assumption violated by van der Pol (cubic term) and, more generally, FitzHugh-Nagumo systems\footnote{In supplementary material, \cite{gorbach2017scalable} consider 2D FitzHugh-Nagumo system but with observed variables, and admit the method does not apply directly so they just fix the state   to a Gaussian Process fit obtained through observations of the state, and do not re-estimate it in their variational framework. However,  this only can work when all state variables are observed, which is  not the case in this work.}.
In summary, the optimization methods discussed above were not immediately applicable to our problem of inferring both  parameters and states of the coupled VDP model, due to local nonlinearity   (i.e., being cubic in each state variable) and presence of unobserved variables. 

As discussed later, given fixed VDP parameters,   state estimation becomes  a nonlinear Kalman smoothing problem \cite{kalman,KalBuc}. 
Optimization-based approaches with nonlinear and non-Gaussian models require iterative optimization techniques;
see for example the survey of~\cite{aravkin2017generalized}.
Dynamical modeling was applied to nonlinear systems early on by~\cite{Anderson:1979,Mortensen1968}.
More recently, the optimization perspective on Kalman smoothing has enabled further extensions, including inference for
systems with unknown parameters~\cite{Bell2000}, systems with constraints~\cite{Bell2009}, and
systems affected by outlier measurements for both linear~\cite{Durovic1999,Meinhold1989,Cipra1997} and
nonlinear~\cite{aravkin2011ell,aravkin2014robust} models.

Our hybrid approach presented later combined variable-projection optimization  with stochastic search, and demonstrated accurate and interpretable results on two calcium imaging datasets.
Note,   however, that a systematic comparison of the proposed approach versus state-of-art on a variety of ODE benchmarks using simulated data (as in most of the work cited above)  remains the direction for future work.

\section{Calcium Imaging Data}

\paragraph{Zebrafish CaI data.}
In~\cite{ahrensNatureMethods2013}, light-sheet microscopy was used to record the neural activity of a whole brain of the larval zebrafish, reported by a genetically-encoded calcium marker, in vivo and at 0.8 Hz sampling rate. 
From the publicly available data~\cite{CaIdata} it is possible to obtain a movie of 500 frames with a 2D collapsed view of 80\% of the approximately 40,000--100,000 neurons in the brain, with a resolution of 400 by 250 pixels (approximately 600 by 270 microns). 
\begin{figure}
\centerline{\includegraphics[width=400pt]{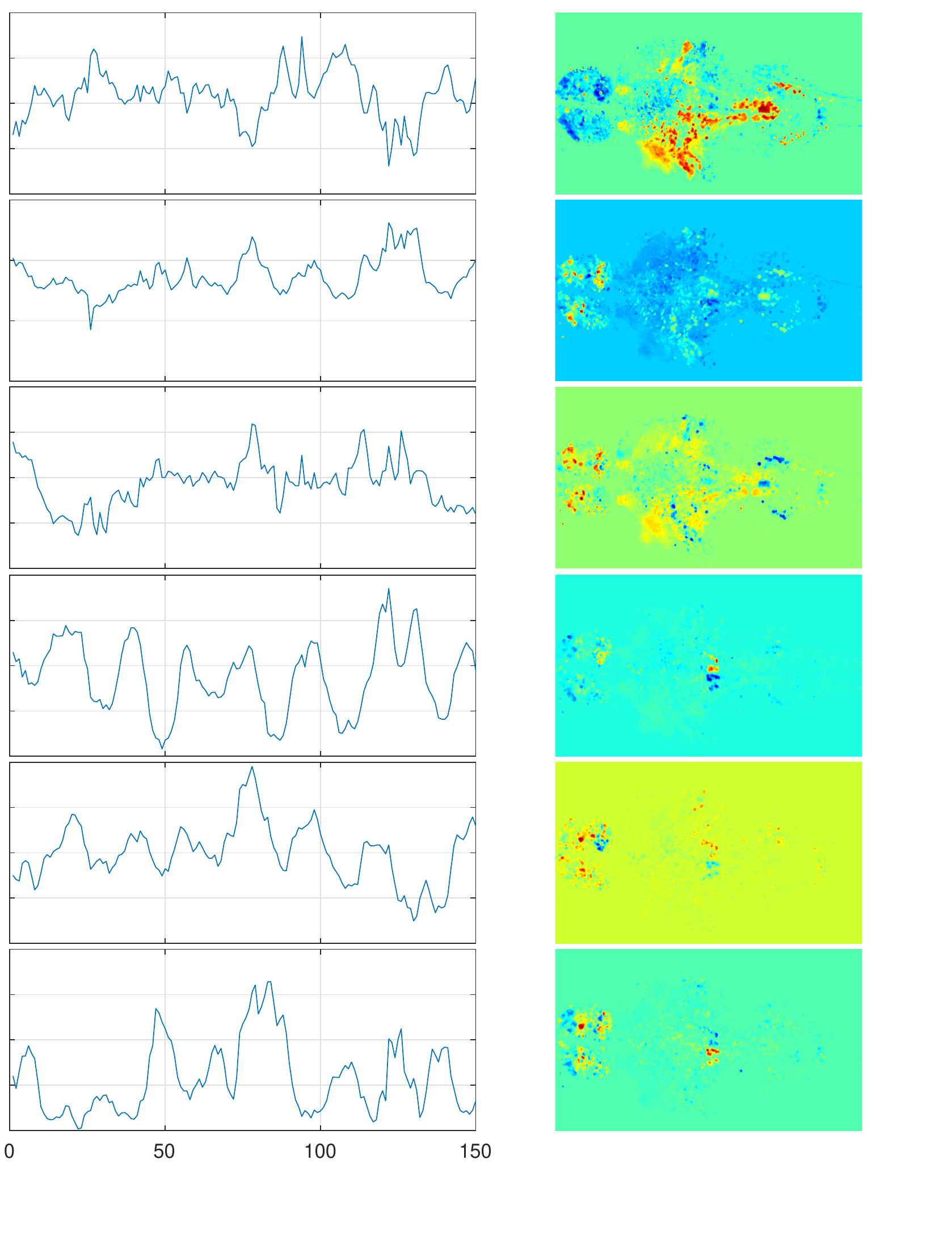} }
\caption{\label{svd} The first 6 SVD components of the zebrafish data.}
\end{figure}
 In order to obtain functionally relevant information, we performed an SVD analysis of these data.
   Figure \ref{svd} shows the first 6 SVD time components (not including the mean, i.e. assuming normalized data) used in the current work.  Left and right columns show the  respective temporal and spacial components. The spatial components show a clear neural substrate, and therefore the time components can be interpreted as traces of neuronal activity from within brain systems identified by each corresponding space components.
For example, spatial components show pronounced but non-overlapping forebrain island-like structures, often with lateral symmetry. Moreover,  spatial components 4-6 include in addition the hindbrain oscillator (seen in the right panels). The corresponding temporal components are dominated by oscillatory activity, consistent with the physiology of the hindbrain oscillator described in~\cite{ahrensNatureMethods2013}.

\paragraph{Rat CaI Data.}
We also considered another calcium imaging dataset, of a rat's visual cortex: conventional cytosolic GCaMP6f data in V1 of a rat looking at oriented gratings. The data was sampled at 44.69 Hz (frames/sec), and later downsampled to approximately match the resolution of the above zebrafish dataset; the experiment was repeated 8 times, yileding 8 datasets. After several preprocessing steps, described in Supplemental Material, we obtained 8 datasets,   over 182 individual neural cells, with  276 time points each.  Top 9 SVD
components were extracted for each dataset (not plotted here due to space limitations).

\section{Van der Pol Model of Neuronal Activity}

Because neuronal calcium dynamics are largely driven by transmembrane voltage and voltage-dependent calcium channels, we model the calcium dynamics of a neuron, or small clusters of them, as a 2D differential equation with a voltage-like variable (activity), and a recovery-like variable (excitability), following similar approaches in the literature~\cite{Izhikevich2007}. Given that one salient feature of neural systems is their propensity for oscillations, as well as sharp transitions from passive to active states, we consider  the following nonlinear oscillator model for each scalar component:
\begin{align}
\label{eq:model}
\dot{x}_{1i}(t) &= \alpha_{1i} x_{1i}(t)(1-{x_{1i}}^2(t)) + \alpha_{2i} x_{2i}(t) + \sum_{j=1}^{m} W_{ij} x_{1j}(t) \nonumber \\
\dot{x}_{2i}(t) &=- x_{1i}(t) 
\end{align}
where $m$ is the number of considered neural units (e.g, SVD components), $x_{1i}(t)$ and $x_{2i}(t)$ represent the (observed) activity and the (hidden) excitability variables of the $i$-th neural unit, respectively, and the $W$ matrix represents the coupling strength between different neural units. Thus,  $I_i(t)=\sum_j W_{ij} x_{1j}(t)$  models the synaptic input to the $i$-th unit provided by other units through their observed $x_{1j}$ variables.
 The parameters $\alpha_{ki}$ 
 determine the bifurcation diagram of the system, allowing for a rich set of dynamical states including oscillations and spike-like responses~\cite{Wiggins2003,Izhikevich2007}. However, imaging techniques only provide information about activity $x_{1i}(t)$, i.e. the calcium concentration in the case of CaI. In consequence, our model-based analysis requires the inference of the excitability variable represented by hidden (unobserved) variables $x_{2i}$.  The above equations represent a bit simplified form of a van der Pol oscillator, where two (linear) terms in the second equation are omitted; still, as we will see, the model is powerful enough to capture the data dynamics.

 When the parameters $\alpha$ and $W$ in~\eqref{eq:model} are known, inferring
the hidden components $x_{2i}(t)$
from observations $x_{1i}(t)$ is a nonlinear Kalman smoothing problem \cite{kalman,KalBuc}. 
We develop here a method to find the hidden variables (${x}_{2i}(t) $) from the observed ones (${x}_{1i}(t)$)
for given parameter settings, and to learn unknown parameter settings themselves.
Indeed, the problems are coupled; however,
rather than using alternating optimization (closely related to EM), we use fast optimization techniques
available for nonlinear Kalman smoothing to fully minimize over the hidden states for each update of the unknown parameters.
The algorithm can be understood in the framework of recent results on variable projection (partial minimization),
which is efficient for dealing with nonconvex, possibly ill-conditioned problems.

While detailed convergence and sensitivity analysis of this algorithm is a topic of ongoing work, 
 {\em this work is the first, to the best of our knowledge,  to propose an approach for learning  a coupled van der Pol oscillator model from data, and evaluate not only data fit but also prediction performance in time series forecasting.}

\section{Learning  van der Pol Model: ODE-Constrained Inference}
\label{sec:Nonlinear}

We discretize the ODE model in equation ~\eqref{eq:model},
and formulate a joint inference problem for the state space $x$
and parameters $\alpha, W$ that is informed by noisy direct observations of some components; and constrained by the discretized dynamics.

\paragraph{Inference for a single component.}
For time index $k$, let  $x_i^k \in \mathbb{R}^2$ denote the $i$th  component of the van der Pol model given earlier in the equation
 ~\eqref{eq:model}, so  $x_i^k = (x^k_{1i},x^k_{2i})^T$, i.e. the state contains both observed and hidden variables.
The discretized dynamics governing the evolution   can be written
\vspace{-0.07in}
\[
x_i^{k+1} = g(x_i^k, \alpha_i),
\]
where $g$ is a first-order Euler discretization of the nonlinear ODE~\eqref{eq:model}.
The $\alpha_i = [\alpha_{1i}, \alpha_{2i}]$ inform the evolution of the entire time series
$x_i = \begin{bmatrix} (x_i^1)^T & \cdots & (x_i^N)^T \end{bmatrix}^T$.
Given an initial and possibly inaccurate state $x_i^0$,
we can describe the dynamics of the entire $i$th component in compact form as
$G(x_i,\alpha) = \eta_i^0$, with
\begin{equation}
\label{ghDef}
G(x_i,\alpha_i) = \begin{bmatrix}x_i^1 \\ x_i^2 - g(x_i^1,\alpha_i) \\ \vdots \\ x_i^N - g(x_i^{N-1},\alpha_i) \end{bmatrix},
\quad
\eta_i^0 = \begin{bmatrix}
x_i^0 \\ 0\\ \vdots \\ 0
\end{bmatrix}.
\end{equation}
Given noisy observations
\[
z_i^k  = H_k x_i^k + \omega_k,
\]
where H is just a linear measurement function (i.e.,  a binary "mask" extracting observed part $x_{1i}$ from  the   i-th component $x_i=(x_{1i},x_{2i})^T$),
we obtain ODE-constrained optimization problem for the $i$th component:
\begin{equation}
\label{eq:formulation}
\min_{x_i,\alpha} \quad \frac{1}{2}\|z_i - Hx_i\|^2 \quad \mbox{s.t.} \quad G(x_i,\alpha_i) = \eta_i^0,
\end{equation}

Problem~\eqref{eq:formulation} is challenging because (1) the ODE constraint function $G$
is nonlinear in $x_i$, and (2) because it is a joint optimization problem over $\alpha_i$ and $x_i$.
To solve this problem, we use the technique of partial minimization~\cite{aravkin2017efficient}\footnote{For
particular instances, partial minimization is often called {\it variable projection}.},
often used in PDE-constrained optimization~\cite{van2015penalty}. Rewriting~\eqref{eq:formulation} with a quadratic penalty, we obtain the relaxed problem
\begin{equation}
\min_{x_i,\alpha_i} f_{\lambda}(x_i,\alpha_i) := \frac{1}{2}\|z_i - Hx_i\|^2 + \frac{\lambda}{2}\|G(x_i,\alpha_i) - \eta_i^0\|^2.
\end{equation}
The key idea is to then use {partial minimization} with respect to $x_i$ at each iteration
of $\alpha_i$ and optimize the value function:
\[
\min_{\alpha_i} \widetilde f_\lambda(\alpha_i) := \min_{x_i}  f_{\lambda}(x_i,\alpha_i).
\]
The intuitive advantages of this method (find the best state estimate for each $\alpha$ regime) are borne
out by theory. In particular, for a large class of models, the objective function $ \tilde{f_\lambda}(\alpha_i)$ is well-behaved
for large $\lambda$, unlike the joint objective $f_{\lambda}(x_i,\alpha_i)$~\cite{aravkin2017efficient}\footnote{The Lipschitz constant of the gradient of $\widetilde f_\lambda(\cdot)$
stays bounded as $\lambda \rightarrow \infty$, which is clearly false for $f_\lambda(\cdot, \cdot)$.}.

Evaluating $f_\lambda (\alpha)$ requires a minimization routine. We compute
gradient and Hessian approximations
\[
\begin{aligned}
\nabla_{x_i} f_\lambda  & = H^T(Hx_i-z_i) + \lambda G_x^T (G(x_i,\alpha_i) - \eta_o^0) \\
\nabla_{x_i}^2 f_\lambda & \approx H^TH + \lambda G_x^TG_x
\end{aligned}
\]
where $G_x = \nabla_x G(x_i,\alpha_i)$.
Evaluating $f_\lambda$ requires obtaining an (approximate) minimizer $\hat x_i$. With $\hat x_i$ in hand,
$\nabla_{\alpha_i} f_\lambda$ can be computed using the formula
\begin{eqnarray}
\label{eq:fgrad}
\nabla_{\alpha_i} f_\lambda(\alpha_i) \approx  \lambda G_\alpha(\hat x_i, \alpha_i) (G( \hat x_i, \alpha_i) - \eta_i^0), \\
\quad G_\alpha = \nabla_\alpha G(x_i, \alpha_i).
\end{eqnarray}
The accuracy of the inner solve in $x_i$ can be increased as the optimization over $\alpha_i$ proceeds. Constraints can also be placed on $\alpha_i$
to eliminate non-physical regimes or to incorporate prior information.

\paragraph{Extension to \emph{m} components.}
In addition to estimating the dynamic parameters $\alpha$, we are also interested in inferring
the connectivity matrix $W$. Extending the model to \emph{m} components,
let $x$ contain $m$ components $x_i$,  so that
in particular $x^k$ contains $x_1^k, \dots, x_m^k$;
and let $\alpha$ contain $m$ parameter sets $\alpha_i$.
We can now write down the full nonlinear process model $G$ as
\begin{equation}
G(x,\alpha, W) = \begin{bmatrix}x^1 \\ x^2 - g(x^1,\alpha, W) \\ \vdots \\ x^N - g(x^{N-1},\alpha, W) \end{bmatrix},
\end{equation}
with $x\in\mathbb{R}^{2mN}$, and the dynamics in the previous section replicated across the $m$ components.
Without the $W$ matrix, this would be $m$ independent models written jointly. The $W$ adds linear coupling
across the components.

The optimization approach for $m$ components is analogous to the single-component case, but includes $m$
components simultaneously, and also infers the coupling matrix $W$:
\[
\min_{x,\alpha, W} f_{\lambda}(x,\alpha, W) := \frac{1}{2}\|z - Hx\|^2 + \frac{\lambda}{2}\|G(x,\alpha, W) - \eta^0\|^2.
\]
Just as for a single component, we optimize this objective using partial minimization in $x$ and working with the value function
\[
\widetilde f_\lambda(\alpha, W) = \min_x  f_{\lambda}(x,\alpha, W).
\]
We optimize over $x$ at each iteration using the Gauss-Newton method. The outer iteration is a fast projected gradient method for minimizing $\widetilde f_\lambda(\alpha, W)$ subject to
simple bound constraints.

\paragraph{Augmenting VP Optimization with Stochastic Search.} Consistently with prior work~\cite{gabor2015robust,rodriguez2006novel}, hybrid stochastic-deterministic methods tend to perform better than a sole local optimization for complex problems. Also, as we observed empirically, finding good initialization for the parameters and hidden states is crucial for the success of  VP optimization. Thus, we decided to  combine   optimization with a stochastic search;  essentially, it  performs a  random walk  in the parameter and hidden state space,   accepting   random steps that improve certain fitness function and discarding those which did not.  Specifically, we used the following fitness function: $f = \min_{i} (c_i + \gamma R^2_i)$,  where $c_i$ and $R^2_i$ are the Pearson correlation and coefficient of determination of the $i$ observable component, respectively, and $\gamma$ is a  weight hyperparameter. We alternate stochastic search with VP optimization, until convergence in $f$, or until the max number of iterations is reached. For details on the algorithm, see Supplemental Material.

\section{Predicting Future Activity: VDP, VAR, LSTM and VDP-LSTM Hybrid}

\begin{small}
\begin{figure*}[t]
\centerline{ \includegraphics[width=\hsize]{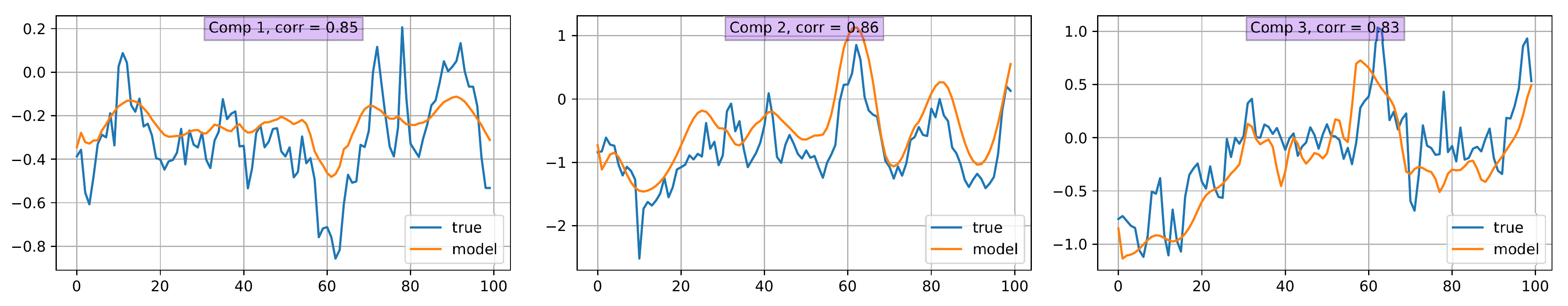}}
 \centerline{(a) Zebrafish CaI}
 \centerline{ \includegraphics[width=\hsize]{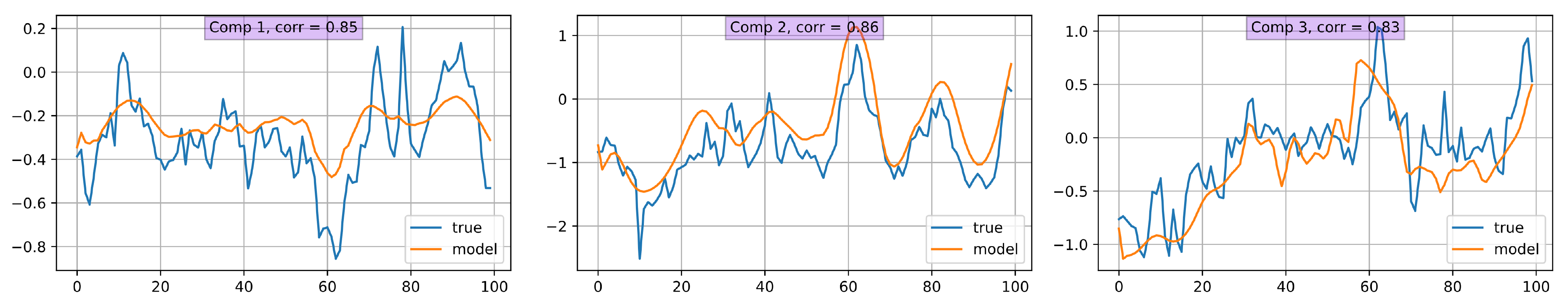}}
 \centerline{(b) Rat CaI}
\caption{\small Van der Pol model fit on 3 top SVD components for (a) zebrafish  and (b) rat calcium imaging datasets 
(each plot shows the Pearson correlations between the true and predicted time series). Each component is   normalized by the mean of the standard deviations from all the components modeled in each case, i.e. 6 for zebrafish and 9 for rat.
}
\label{fig:fit1}
\vspace{-0.2in}
\end{figure*}
\end{small}

\paragraph{VDP}
Given a VDP model (parameters and initial hidden states) estimated on training time interval, we can integrate it not only to fit the training time series, but also to predict the immediate future time points, immediately following the training  interval, since the  hidden state variable at the last training time point is also estimated. Note, however, that the model is not trivially applicable for predicting an arbitrary interval in the remote future, since its hidden state will be unknown.

\paragraph{VAR.}
The simplest baseline for time-series forecasting is the standard linear Vector Auto-Regressive (VAR) model. At a given time $t$, given a $x \times k$ matrix of  $k$ previous observations  ($x_{1,:}^{t-k},..., x_{1,:}^{t-2}, x_{1,:}^{t-1} $) over   $m$ observed temporal components (as mentioned above, $m=6$ and $m=9$ for zebrafish and for rat CaI data, respectively),  
the task is to (jointly) predict      $x_{1,:}^{t}$ at the next time step $t$. To predict  multiple steps ahead, we  proceed recursively, using the previous predictions as new inputs.   we experimented with several values of $k$, and selected   $k=6$, for which  VAR was producing its best results. 
 
\paragraph{LSTM.} We also compared VDP model prediction with several standard time-series forecasting approaches, including recurrent neural networks, such as popular LSTM  model \cite{19-Hochreiter1997}. We used a two-layer LSTM network, trained in Keras using RMSprop with dropout; see Supplemental Material for details on network's architecture and hyperparameters. Note that, to be fair to the linear VAR model, we provided LASTM with the same $k=6$ previous time points, and predicted multiple steps ahead as discussed above for VAR.
 
\paragraph{Novel Approach: VdP-LSTM (LSTM Pretrained with vdP Simulations).}
Training LSTMs often requires a large number of training samples, which was not the case in our data, so some data-augmentation could help. A simple, and often used, approach is to generate noisy versions of the training data. However, a more effective alternative, as we show below,    was to use VDP model as a possible  ''ground-truth" dynamical model for simulating (unlimited) additional training data. 
The resulting hybrid approach, where VDP is first estimated on training data, and then used for pretraining LSTM, followed by fine-tuning LSTM on real training data, is called here {\em VDP-LSTM}.

 Pretraining on VDP-simulated data can serve as a  regularizer (prior) in the absence of large training data sets; note that estimating VDP from smaller data might be easier than training LSTM since VDP has much less parameters to train, but hopefully captures some domain knowledge, e.g. certain characteristics of neuronal activation dynamics, while LSTM is a general-purpose model. For implementation details, see Supplemental Material.

\section{Empirical Evaluation}


\begin{figure*}[t]
\begin{multicols}{4}
    \hspace{-0.15in}
    \includegraphics[width=1.53in,height=1.5in]{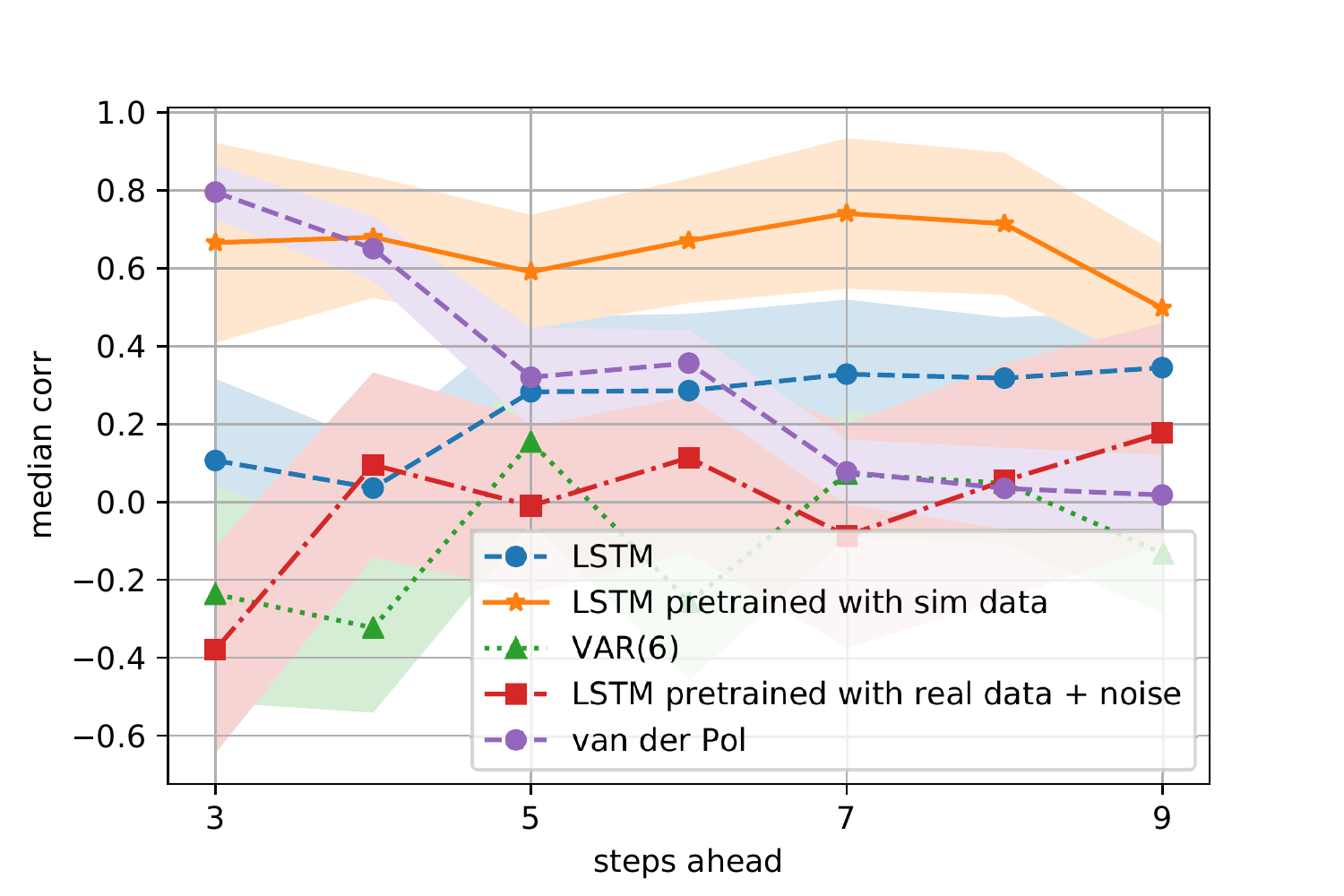}
    \par \vspace{-0.1in}\caption{ \small  Zebrafish: correlation, short-term.
    }
    \label{fig:fish_corr}
  \hspace{-0.2in} 
      \includegraphics[width=1.53in,height=1.5in]{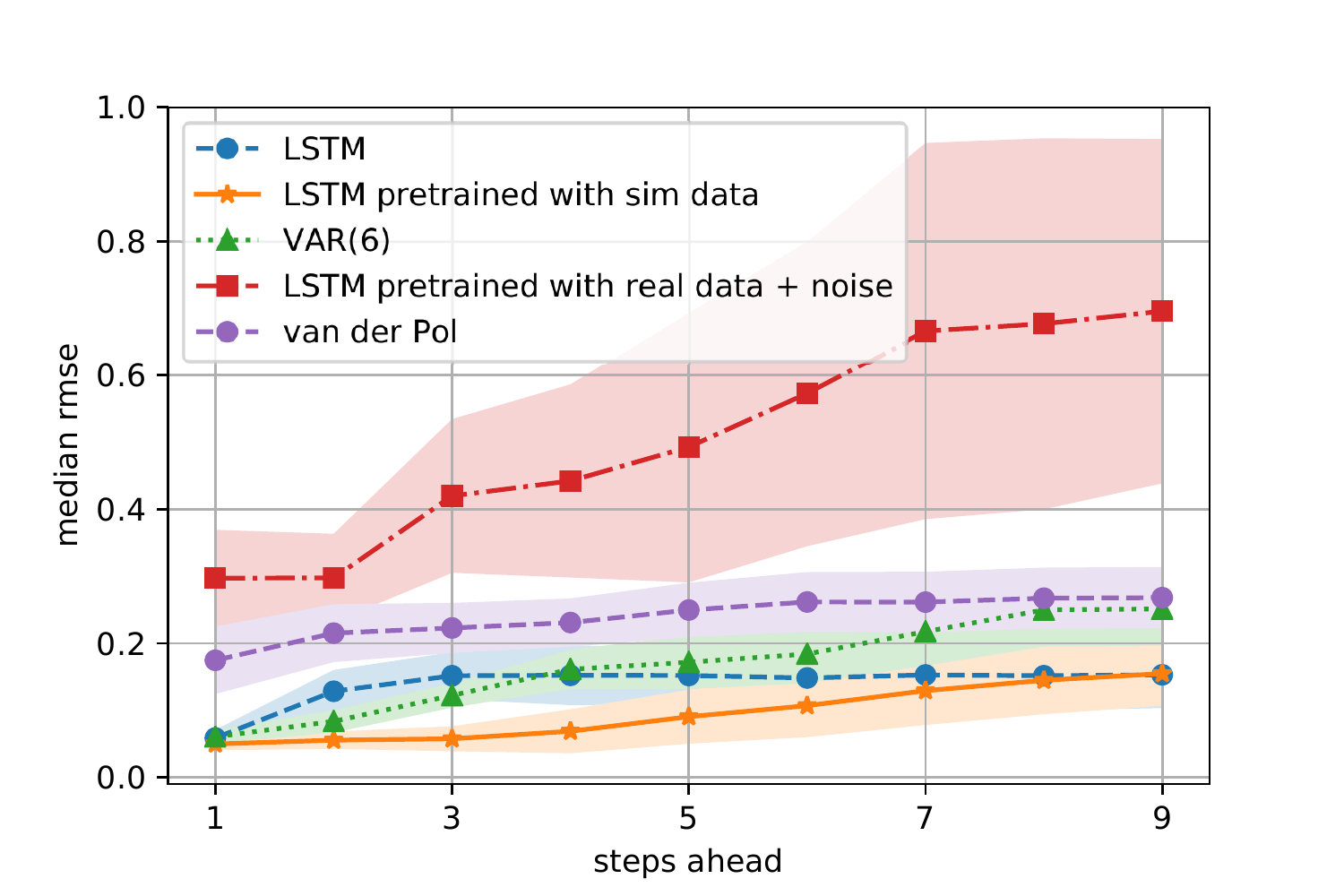}
    \par \vspace{-0.1in}\caption{ \small 
    Zebrafish: RMSE, short-term.
     }
  \label{fig:fish_rmse}
  \hspace{-0.2in}
     \includegraphics[width=1.53in,height=1.5in]{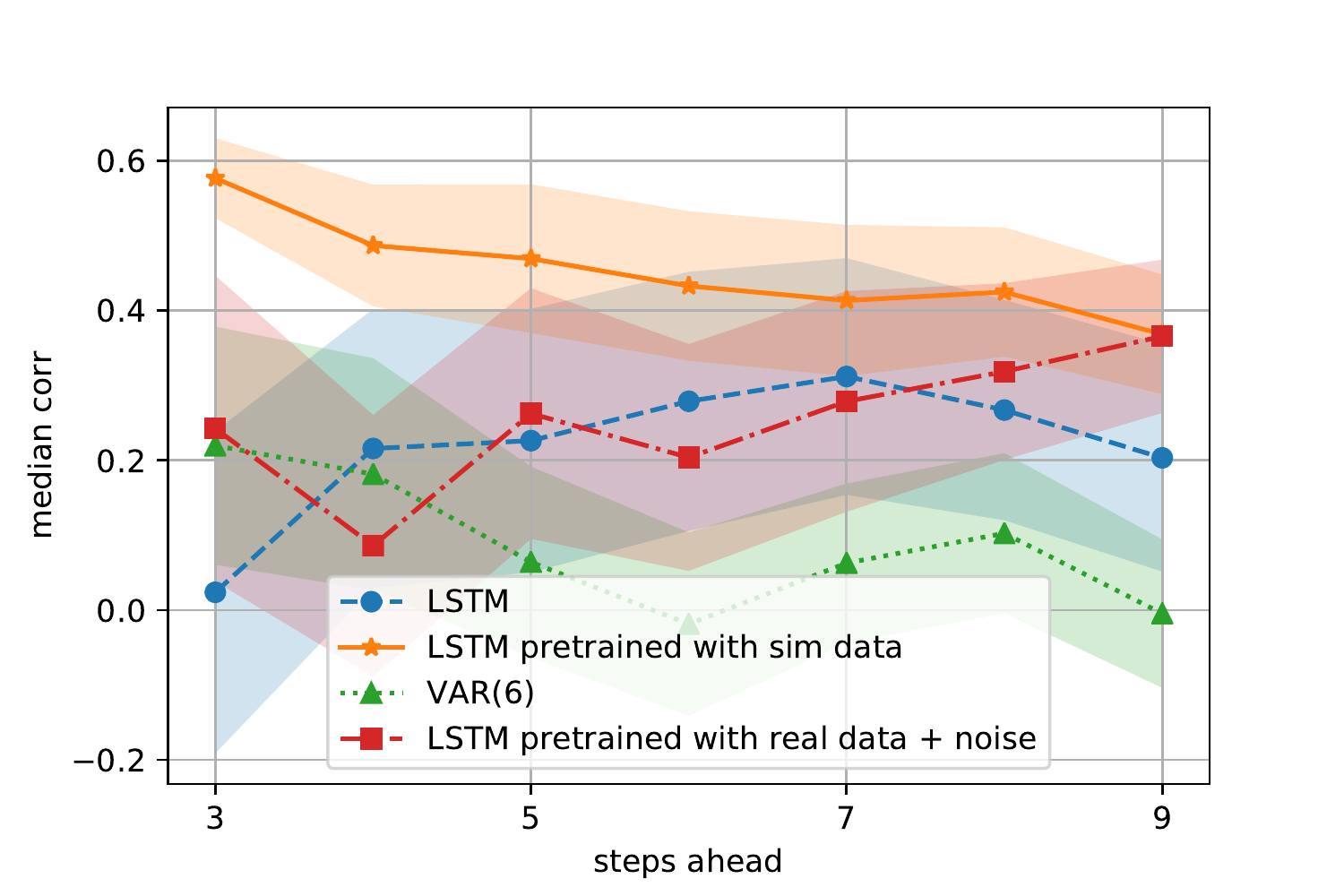}
    \par \vspace{-0.1in}\caption{ \small Zebrafish: correlation,long-term. 
    }\label{fig:fish_corr1}
      \hspace{-0.2in}
     \includegraphics[width=1.53in,height=1.5in]{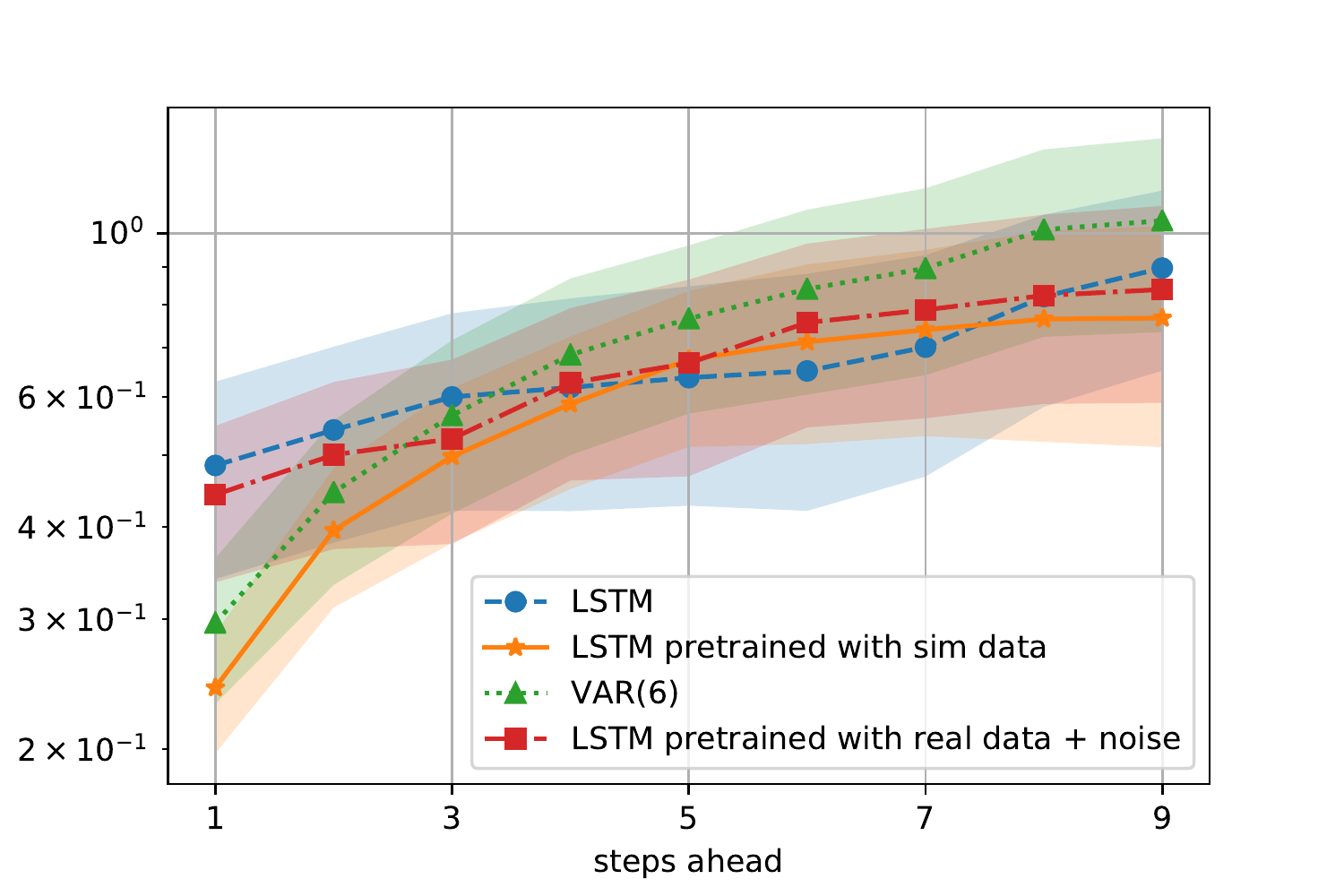}
    \par \vspace{-0.1in}\caption{ \small Zebrafish:  RMSE, long-term.
    }\label{fig:fish_rmse1}
\end{multicols}
\vspace{-0.35in}
\begin{multicols}{4}
    \hspace{-0.15in}
    \includegraphics[width=1.53in,height=1.5in]{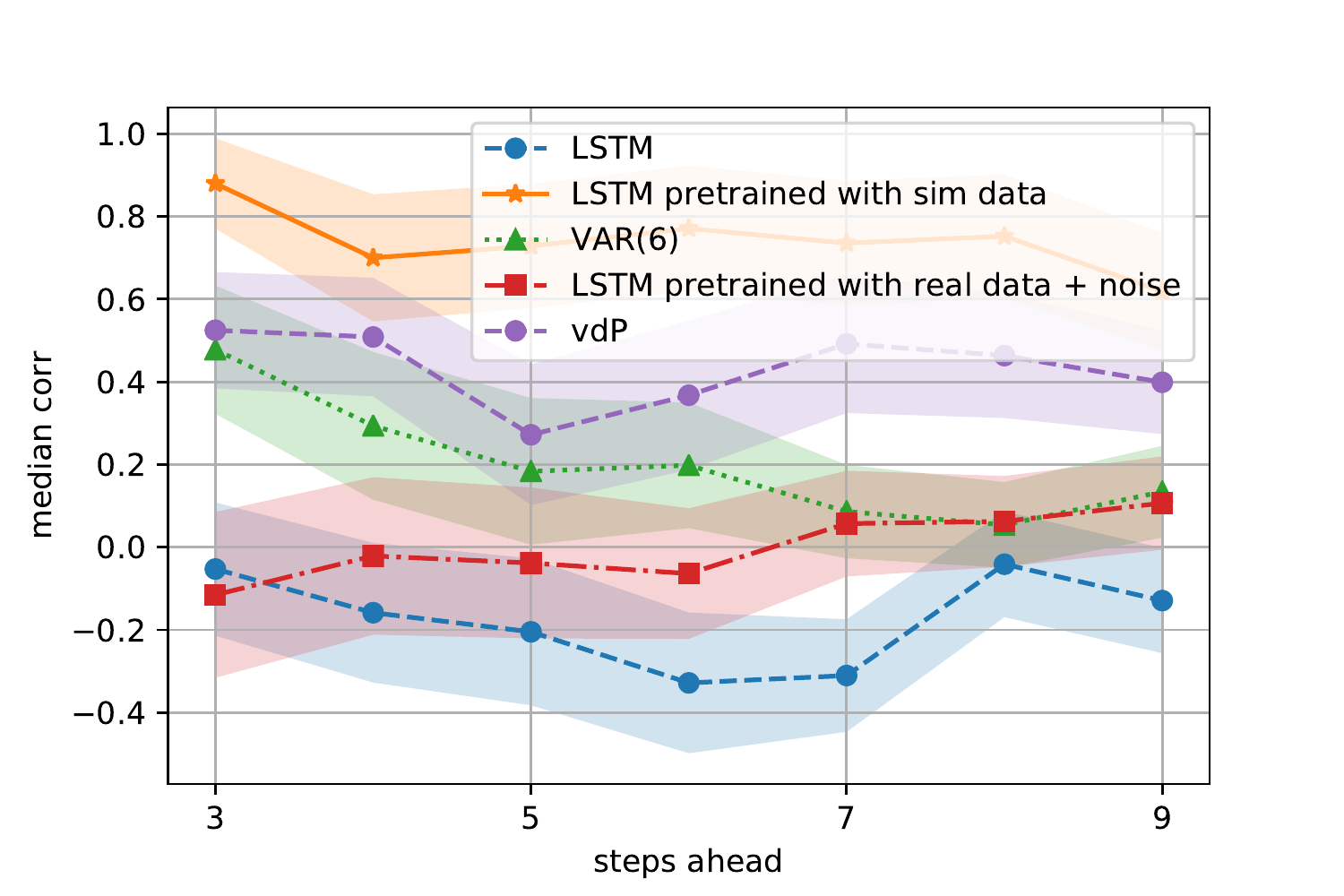}
    \par \vspace{-0.1in}\caption{ \small  Rat:  correlation, short-term.
    }
    \label{fig:rat_corr}
  \hspace{-0.2in} 
      \includegraphics[width=1.53in,height=1.5in]{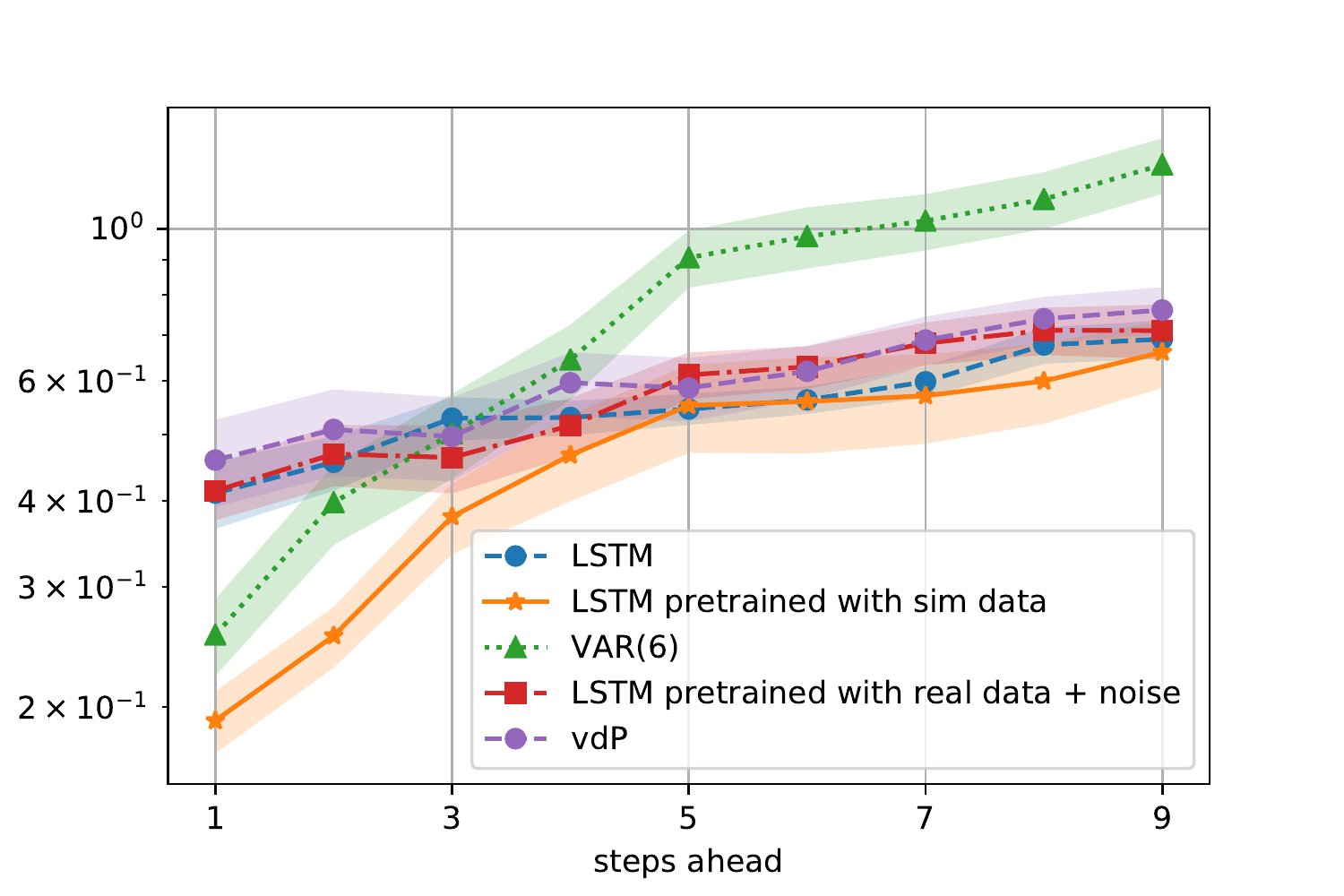}
    \par \vspace{-0.1in}\caption{ \small 
    Rat:  RMSE, short-term.
    }
  \label{fig:rat_rmse}
  \hspace{-0.2in}
     \includegraphics[width=1.53in,height=1.5in]{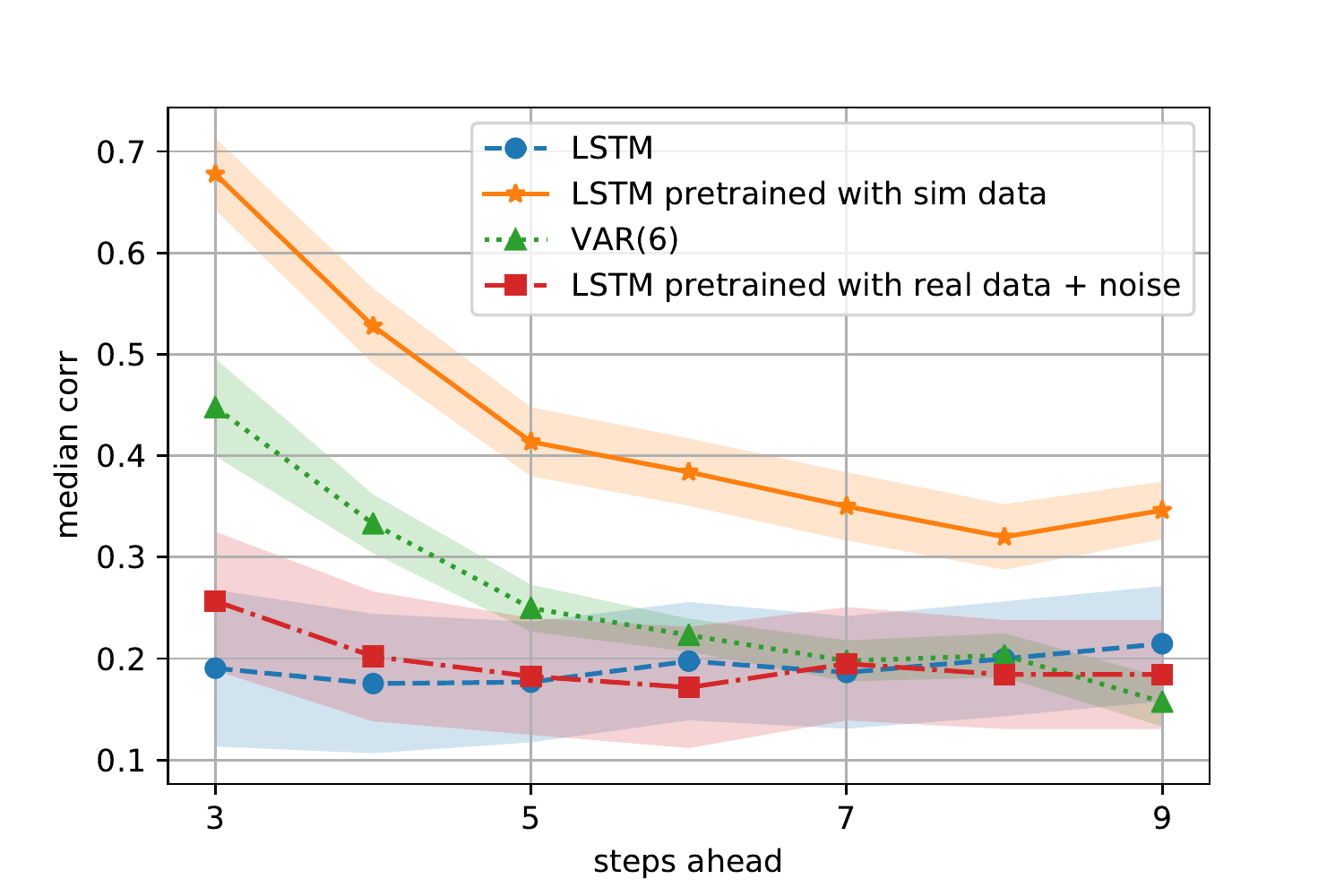}
    \par \vspace{-0.1in}\caption{ \small Rat,   correlation, long-term.
    }\label{rat:fish_corr1}
  \hspace{-0.2in}
 \includegraphics[width=1.53in,height=1.5in]{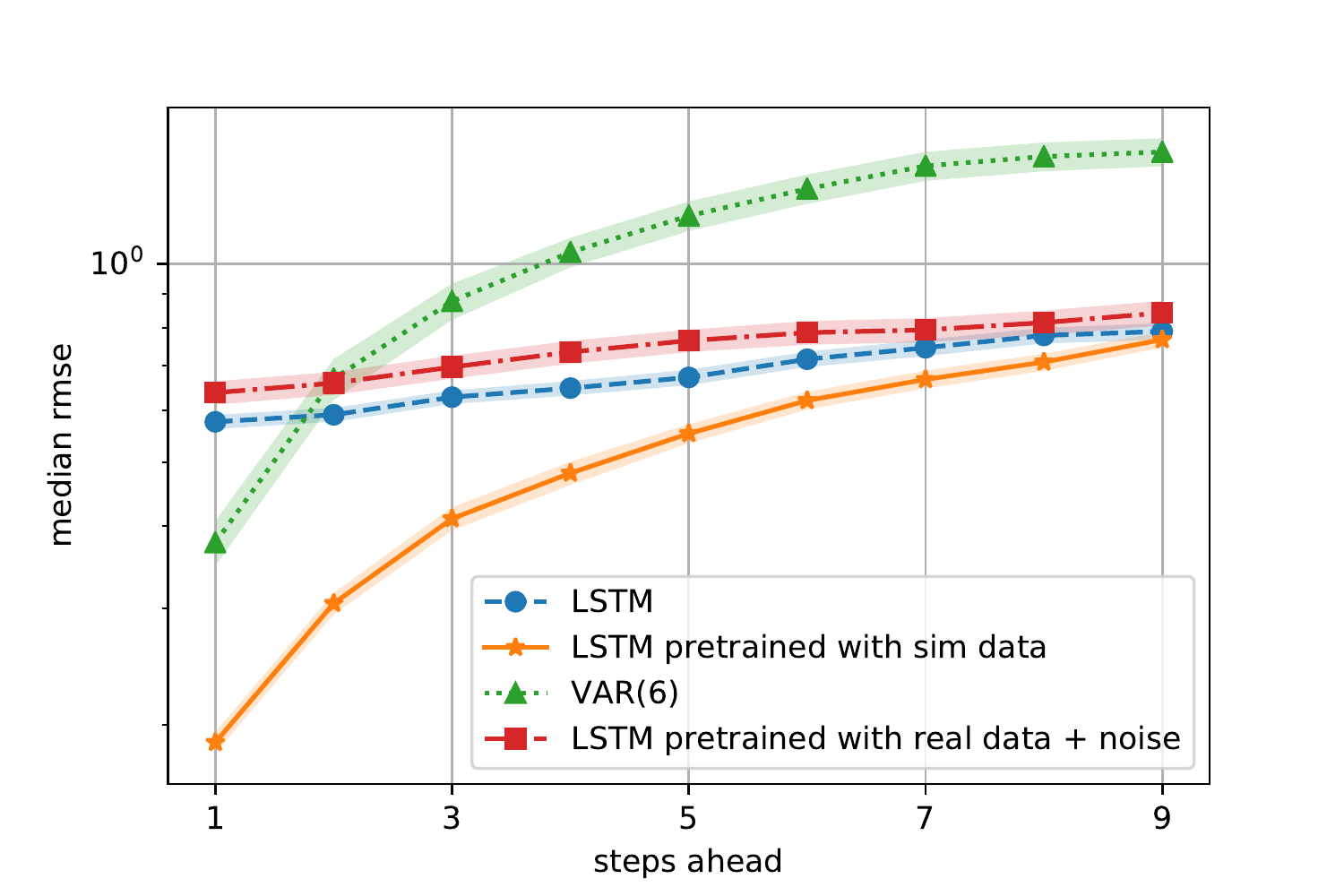}\par \vspace{-0.1in}\caption{\small Rat:   RMSE, long-term. }
   \hspace{0.15in}
  \label{fig:rat_rmse1}
\end{multicols}
\vspace{-0.4in}
\end{figure*}

We applied the proposed approach(es)  on two calcium imaging datasets discussed earlier, zebrafish (whole-brain)  and rat's visual cortex calcium imaging,  evaluating the following key aspects:
  (1)   VDP model fit on training data; (2) predictive accuracy when forecasting time series using VDP, LSTM and hybrid VDP-LSTM, with a linear Vector Auto-Regressive (VAR) model used as a baseline, and another baseline, representing LSTM pretrained with a noisy version of training data (see Supplemental Material for details); (3)  interpreting VDP's coupling (interaction) matrix.  
  Zebrafish data were split into 5 training segments, 100 time points each; the next  20 time points for each training segment were used for prediction/testing.
Each of 8 rat datasets was used as a segment above, split into 100 training and 176 testing points.

\paragraph{Evaluating VDP model fit.}
We evaluated multiple runs of VDP estimation procedure described above, combining stochastic search with VP optimization.
Figure \ref{fig:fit1} shows the fit to the first 3 SVD components from each training dataset, achieved by one of the best-performing models; note the correlations between the actual data and the model predictions are high, ranging from 0.83 to 0.94 (similarly high correlations from 0.82 to 0.94 were achieved for the remaining SVD components, as shown in Supplemental material).

\begin{figure}
 \centerline{\includegraphics[width=0.5\columnwidth]{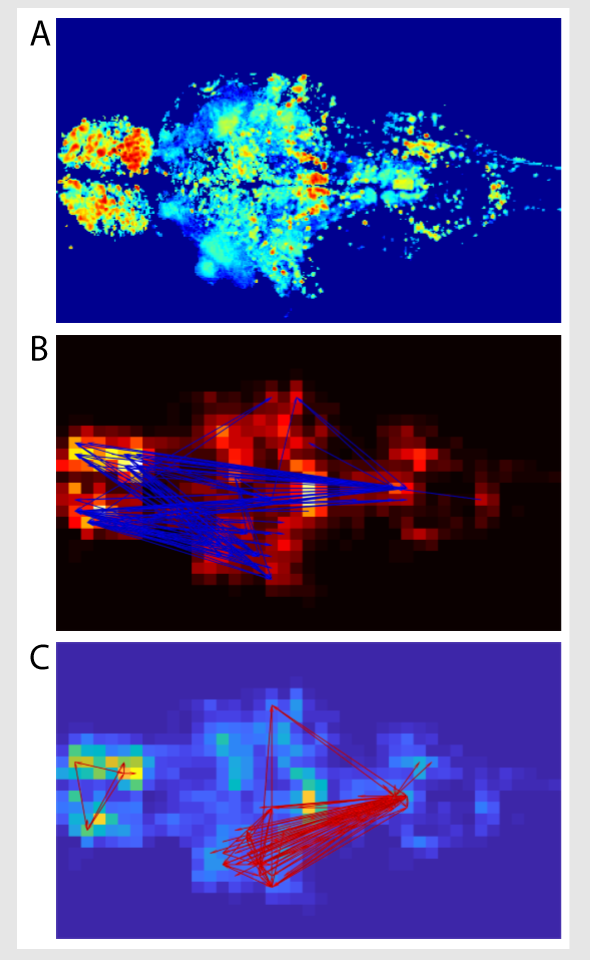}}
 \caption{ \label{fig:anatomy} A. Sum of all spatial components reveals anatomical regions participating in the network of components model. B. Identified dominant pixel-to-pixel edges   visualized as directed links (blue) for excitatory (consistently strongly positive) connections. Note that excitation predicted by this analysis joins midline hindbrain motor nuclei to bilateral forebrain areas, and reciprocally joins forebrain motor areas to unilateral tectum and pretectum. C. Identified dominant pixel-to-pixel edges for inhibitory (consistently strongly negative) connections. Predicted inhibition joins a sparse set of forebrain pixels in a bidirectionally coupled loop. Strong reciprocal inhibition identified also between unilateral tectum and midline hindbrain motor nuclei.}
 \end{figure}

\paragraph{ Interpretability.}
In Figure \ref{fig:anatomy}, we demonstrate an example of connectivity analysis with respect to the known anatomy of zebrafish. First, we serialized each spatial
component  as a vector, and computed  vector product of each spatial component combination,  scaled by the product of the square roots of each corresponding eigenvalue and the resulting matrix of pixel-to-pixel base connection strengths were stored. 

 To compute the final pixel to pixel weight matrix, we summed the  weight  matrices for all VDP models generated   as described  before. 
The element of the summed matrix were each then multiplied by the corresponding pixel-to-pixel base connection strength matrix, and the resulting 
weighted matrices were then summed to create the final pixel-to-pixel weight matrix. For visualization only, the top 200 positive weights were rendered as pixel to pixel  excitatory edges (blue) and the lowest negative 200 as inhibitory edges (red).

The results show connections between brain areas that are consistent with known anatomical demarcations in the larval zebra fish brain \cite{ahrensNatureMethods2013} as well as functional relationships between forebrain, midbrain, and hindbrain. The lateralization of certain connections are intriguing and warrant further analysis, as does the apparent involvement of both visual areas and motor generating areas in the final network. Finally, a sparse competitive recurrent network is implicated for the fish forebrain intraconnectivity.

\paragraph{ Predicting future activity.} Figures \ref{fig:fish_corr} to \ref{fig:rat_rmse1} compare the performance of several predictive methods: vector autoregressive (VAR) model (green), van der Pol model (violet), LSTM (blue), and vdP-LSTM, i.e. LSTM pretrained on the data simulated using the above van der Pol model (orange), as well as LSTM pretrained using noisy version of training data (red).

As mentioned above, we train the models  on 100 consecutive points of each data segment (5 for zebrafish and 8 for rat), and then used the next 20 points (zebrafish) or next 176 points (rat) for prediction.  First, we predict the  next 9 points (x-axis plots the index of the time points being predicted) immediately following the training interval ( {\em short-term} prediction). Then, we  {\em long-term prediction} is performed by moving forward, by 1 time point,  the 9-point window, until we reach the end of the  test dataset. Each time we update  VAR and LSTM models,  using  the last 6 points preceding that 9-point test interval, as described earlier. As noted above, we could not evaluate VDP for long-term prediction without adding a mechanism for inferring its hidden states, not readily available for future points beyond the last training point; thus, VDP plot is  omitted from the corresponding long-term comparisons.

We compute the  median Pearson correlation between the true and predicted time series, and the median root-means square error (RMSE),  over all training/test splits (i.e., 5    for  zebrafish and 8 for rat), all SVD components and all testing intervals (in case of short-term, we only had 1 test interval). Shaded area around each curve represents the standard error. Note that, for long-term prediction, having 176 instead of 20 test samples  yields considerable lower variance (narrower error bars) for rat as compared to zebrafish.

The results are quite consistent across datasets and evaluation types: linear VAR model (green) performs poorly, unable to capture the nonlinear dynamics; van der Pol (violet) outperforms LSTM (blue) in the beginning, but then LSTM catches up; the hybrid vdP-LSTM model combines the best of both.
Similarly, the hybrid approach performs best in terms of RMSE error.

\section{Conclusions}
Motivated by the challenging problem of modeling nonlinear dynamics of brain activations in calcium imaging, we propose a new  approach for learning a nonlinear differential equation model: a variable-projection optimization approach to estimate the parameters of the multivariate coupled van der Pol oscillator. We show how to learn this nonlinear dynamical model,  and demonstrate that it can accurately capture   nonlinear dynamics of the brain data. Furthermore, in order to improve the predictive accuracy when forecasting future brain activity, we used the learned van der Pol to pretrain LSTM networks,  thus imposing an oscillator prior on LSTM; the resulting approach  achieves highest predictive accuracy   among all methods we evaluated.

\bibliography{biblio}

\begin{thebibliography}{10}

\bibitem{ahrensNature2012}
MB~Ahrens et~al.
\newblock Brain-wide neuronal dynamics during motor adaptation in zebrafish.
\newblock {\em Nature}, 485:471--477, 2012.

\bibitem{CaIdata}
MB~Ahrens et~al.
\newblock {Discover Magazine Youtube Channel}.
  \url{https://youtu.be/lppAwkek6DI}, March 2013.

\bibitem{ahrensNatureMethods2013}
MB~Ahrens et~al.
\newblock Whole-brain functional imaging at cellular resolution using
  light-sheet microscopy.
\newblock {\em Nature Methods}, 10:413--420, 2013.

\bibitem{alonso2019}
Solovey G. Yanagawa T. Proekt A. Cecchi G. A. Magnasco M.~O. Alonso, L.~M.
\newblock Single-trial classification of awareness state during anesthesia by
  measuring critical dynamics of global brain activity.
\newblock {\em Scientific reports}, 9(1):4927, 2019.

\bibitem{Anderson:1979}
B.~D.~O. Anderson and J.~B. Moore.
\newblock {\em Optimal Filtering}.
\newblock Prentice-Hall, Englewood Cliffs, N.J., USA, 1979.

\bibitem{Apthorpe_NIPS2016}
Noah Apthorpe, Alexander Riordan, Robert Aguilar, Jan Homann, Yi~Gu, David
  Tank, and H.~Sebastian Seung.
\newblock Automatic neuron detection in calcium imaging data using
  convolutional networks.
\newblock In {\em Advances in Neural Information Processing Systems
  (NIPS-2016)}, pages 3270--3278. 2016.

\bibitem{aravkin2017generalized}
Aleksandr Aravkin, James~V Burke, Lennart Ljung, Aurelie Lozano, and Gianluigi
  Pillonetto.
\newblock Generalized kalman smoothing: Modeling and algorithms.
\newblock {\em Automatica}, 86:63--86, 2017.

\bibitem{aravkin2011ell}
Aleksandr~Y Aravkin, Bradley~M Bell, James~V Burke, and Gianluigi Pillonetto.
\newblock An l1-laplace robust kalman smoother.
\newblock {\em IEEE Transactions on Automatic Control}, 56(12):2898--2911,
  2011.

\bibitem{aravkin2014robust}
Aleksandr~Y Aravkin, James~V Burke, and Gianluigi Pillonetto.
\newblock Robust and trend-following student's t kalman smoothers.
\newblock {\em SIAM Journal on Control and Optimization}, 52(5):2891--2916,
  2014.

\bibitem{aravkin2017efficient}
Aleksandr~Y Aravkin, Dmitriy Drusvyatskiy, and Tristan van Leeuwen.
\newblock Efficient quadratic penalization through the partial minimization
  technique.
\newblock {\em IEEE Transactions on Automatic Control}, 2017.

\bibitem{babtie2014topological}
Ann~C Babtie, Paul Kirk, and Michael~PH Stumpf.
\newblock Topological sensitivity analysis for systems biology.
\newblock {\em Proceedings of the National Academy of Sciences},
  111(52):18507--18512, 2014.

\bibitem{Bell2009}
B.~M. Bell, J.~V. Burke, and G.~Pillonetto.
\newblock An inequality constrained nonlinear {Kalman-Bucy} smoother by
  interior point likelihood maximization.
\newblock {\em Automatica}, 45(1):25--33, 2009.

\bibitem{Bell2000}
B.M. Bell.
\newblock The marginal likelihood for parameters in a discrete {Gauss-Markov}
  process.
\newblock {\em IEEE Transactions on Signal Processing}, 48(3):870--873, 2000.

\bibitem{calderhead2009accelerating}
Ben Calderhead, Mark Girolami, and Neil~D Lawrence.
\newblock Accelerating bayesian inference over nonlinear differential equations
  with gaussian processes.
\newblock In {\em Advances in neural information processing systems}, pages
  217--224, 2009.

\bibitem{Cipra1997}
T.~Cipra and R.~Romera.
\newblock {Kalman} filter with outliers and missing observations.
\newblock {\em Sociedad de Estadistica e Invastigacion Operativa},
  6(2):379--395, 1997.

\bibitem{Feinberg2010}
Feinberg DA and et~al.
\newblock Multiplexed echo planar imaging for sub-second whole brain fmri and
  fast diffusion imaging.
\newblock {\em PLoS ONE}, 5(12):e15710.

\bibitem{destexhe2009}
T~Destexhe~A., Sejnowski.
\newblock The wilson–cowan model, 36 years later.
\newblock {\em Biological cybernetics}, 101(1):1--2, 2009.

\bibitem{dondelinger2013ode}
Frank Dondelinger, Dirk Husmeier, Simon Rogers, and Maurizio Filippone.
\newblock Ode parameter inference using adaptive gradient matching with
  gaussian processes.
\newblock In {\em Artificial intelligence and statistics}, pages 216--228,
  2013.

\bibitem{Durovic1999}
Z.~M. Durovic and B.~D. Kovachevic.
\newblock Robust estimation with unknown noise statistics.
\newblock {\em IEEE Transactions on Automatic Control}, 44(6):1292--1296, June
  1999.

\bibitem{peterlinPNAS2000}
Peterlin et~al.
\newblock Optical probing of neuronal circuits with calcium indicators.
\newblock {\em PNAS}, 97(7):3619--3624, 2000.

\bibitem{takemuraNature2013}
Takemura et~al.
\newblock A visual motion detection circuit suggested by drosophila
  connectomics.
\newblock {\em Nature}, 500:175--181, 2013.

\bibitem{fitzhugh1961impulses}
Richard FitzHugh.
\newblock Impulses and physiological states in theoretical models of nerve
  membrane.
\newblock {\em Biophysical journal}, 1(6):445--466, 1961.

\bibitem{gabor2015robust}
Attila G{\'a}bor and Julio~R Banga.
\newblock Robust and efficient parameter estimation in dynamic models of
  biological systems.
\newblock {\em BMC systems biology}, 9(1):74, 2015.

\bibitem{giovannucci2017onacid}
Andrea Giovannucci, Johannes Friedrich, Matt Kaufman, Anne Churchland, Dmitri
  Chklovskii, Liam Paninski, and Eftychios~A Pnevmatikakis.
\newblock Onacid: Online analysis of calcium imaging data in real time.
\newblock In {\em Advances in Neural Information Processing Systems}, pages
  2381--2391, 2017.

\bibitem{gorbach2017scalable}
Nico~S Gorbach, Stefan Bauer, and Joachim~M Buhmann.
\newblock Scalable variational inference for dynamical systems.
\newblock In {\em Advances in Neural Information Processing Systems}, pages
  4806--4815, 2017.

\bibitem{guckenheimer2013nonlinear}
John Guckenheimer and Philip Holmes.
\newblock {\em Nonlinear oscillations, dynamical systems, and bifurcations of
  vector fields}, volume~42.
\newblock Springer Science \& Business Media, 2013.

\bibitem{havlicek2011dynamic}
Martin Havlicek, Karl~J Friston, Jiri Jan, Milan Brazdil, and Vince~D Calhoun.
\newblock Dynamic modeling of neuronal responses in fmri using cubature kalman
  filtering.
\newblock {\em Neuroimage}, 56(4):2109--2128, 2011.

\bibitem{19-Hochreiter1997}
Sepp Hochreiter and J{\"{u}}rgen Schmidhuber.
\newblock {Long Short-Term Memory}.
\newblock {\em Neural Computation}, 9(8):1735--1780, 1997.

\bibitem{robust-estimation-of-neural-signals-in-calcium-imaging}
Hakan Inan, Murat Erdogdu, and Mark Schnitzer.
\newblock Robust estimation of neural signals in calcium imaging.
\newblock In {\em Neural Information Processing Systems (NIPS-2017)}. 2017.

\bibitem{Izhikevich2007}
Eugene~M. Izhikevich.
\newblock {\em Dynamical Systems in Neuroscience: The Geometry of Excitability
  and Bursting}.
\newblock MIT Press,, 2007.

\bibitem{kalman}
R.~E. Kalman.
\newblock A new approach to linear filtering and prediction problems.
\newblock {\em Transactions of the AMSE - Journal of Basic Engineering},
  82(D):35--45, 1960.

\bibitem{KalBuc}
R.~E. Kalman and R.~S. Bucy.
\newblock New results in linear filtering and prediction theory.
\newblock {\em Trans. ASME J. Basic Eng}, 83:95--108, 1961.

\bibitem{kawahara1980}
T.~Kawahara.
\newblock Coupled van der pol oscillators—a model of excitatory and
  inhibitory neural interactions.
\newblock {\em Biological cybernetics}, 39(1):37--43, 1980.

\bibitem{korn2003there}
Henri Korn and Philippe Faure.
\newblock Is there chaos in the brain? ii. experimental evidence and related
  models.
\newblock {\em Comptes rendus biologies}, 326(9):787--840, 2003.

\bibitem{macdonald2015gradient}
Benn Macdonald and Dirk Husmeier.
\newblock Gradient matching methods for computational inference in mechanistic
  models for systems biology: a review and comparative analysis.
\newblock {\em Frontiers in bioengineering and biotechnology}, 3:180, 2015.

\bibitem{Meinhold1989}
R.~J. Meinhold and N.~D. Singpurwalla.
\newblock Robustification of {Kalman} filter models.
\newblock {\em Journal of the American Statistical Association},
  84(406):479--486, June 1989.

\bibitem{moirogiannis2017renormalization}
Dimitrios Moirogiannis, Oreste Piro, and Marcelo~O Magnasco.
\newblock Renormalization of collective modes in large-scale neural dynamics.
\newblock {\em Journal of Statistical Physics}, 167(3-4):543--558, 2017.

\bibitem{Mortensen1968}
R.E. Mortensen.
\newblock Maximum-likelihood recursive nonlinear filtering.
\newblock {\em Journal of Optimization Theory and Applications}, 2(6):386--394,
  1968.

\bibitem{quach2007estimating}
Minh Quach, Nicolas Brunel, and Florence d'Alch{\'e} Buc.
\newblock Estimating parameters and hidden variables in non-linear state-space
  models based on odes for biological networks inference.
\newblock {\em Bioinformatics}, 23(23):3209--3216, 2007.

\bibitem{ramsay2007parameter}
Jim~O Ramsay, Giles Hooker, David Campbell, and Jiguo Cao.
\newblock Parameter estimation for differential equations: a generalized
  smoothing approach.
\newblock {\em Journal of the Royal Statistical Society: Series B (Statistical
  Methodology)}, 69(5):741--796, 2007.

\bibitem{rodriguez2006novel}
Maria Rodriguez-Fernandez, Jose~A Egea, and Julio~R Banga.
\newblock Novel metaheuristic for parameter estimation in nonlinear dynamic
  biological systems.
\newblock {\em BMC bioinformatics}, 7(1):483, 2006.

\bibitem{Schwartz2014}
DA~Schwartz and et~al.
\newblock Chronic, wireless recordings of large-scale brain activity in freely
  moving rhesus monkeys.
\newblock {\em Nature Methods (advance online pub.)}, 2014.

\bibitem{seung2000}
Lee D. D. Reis B. Y. Tank D.~W. Seung, H.~S.
\newblock Stability of the memory of eye position in a recurrent network of
  conductance-based model neurons.
\newblock {\em Neuron}, 26(1):259--271, 2009.

\bibitem{sitz2002estimation}
Andre Sitz, Udo Schwarz, J{\"u}rgen Kurths, and Henning~U Voss.
\newblock Estimation of parameters and unobserved components for nonlinear
  systems from noisy time series.
\newblock {\em Physical review E}, 66(1):016210, 2002.

\bibitem{Speiser_NIPS2017}
Artur Speiser, Jinyao Yan, Evan~W Archer, Lars Buesing, Srinivas~C Turaga, and
  Jakob~H Macke.
\newblock Fast amortized inference of neural activity from calcium imaging data
  with variational autoencoders.
\newblock In {\em Neural Information Processing Systems (NIPS-2017)}, pages
  4024--4034. 2017.

\bibitem{van2015penalty}
Tristan van Leeuwen and Felix~J Herrmann.
\newblock A penalty method for pde-constrained optimization in inverse
  problems.
\newblock {\em Inverse Problems}, 32(1):015007, 2015.

\bibitem{varah1982spline}
James~M Varah.
\newblock A spline least squares method for numerical parameter estimation in
  differential equations.
\newblock {\em SIAM Journal on Scientific and Statistical Computing},
  3(1):28--46, 1982.

\bibitem{voss2004nonlinear}
Henning~U Voss, Jens Timmer, and J{\"u}rgen Kurths.
\newblock Nonlinear dynamical system identification from uncertain and indirect
  measurements.
\newblock {\em International Journal of Bifurcation and Chaos},
  14(06):1905--1933, 2004.

\bibitem{Wiggins2003}
S.~Wiggins.
\newblock {\em Introduction to Applied Nonlinear Dynamical Systems and Chaos}.
\newblock Springer, 2003.

\end{thebibliography}
\bibliographystyle{plain}

\end{document}